\documentclass[sn-nature]{sn-jnl}

\usepackage{natbib}
\usepackage{graphicx}%
\usepackage{multirow}%
\usepackage{amsmath,amssymb,amsfonts}%
\usepackage{amsthm}%
\usepackage{mathrsfs}%
\usepackage[title]{appendix}%
\usepackage{xcolor}%
\usepackage{textcomp}%
\usepackage{manyfoot}%
\usepackage{booktabs}%
\usepackage{algorithm}%
\usepackage{algorithmicx}%
\usepackage{algpseudocode}%
\usepackage{listings}%
\usepackage{csquotes}

\usepackage{multirow}
\usepackage{array}

\let\cline\cmidrule 

\theoremstyle{thmstyleone}%
\theoremstyle{thmstyletwo}%

\theoremstyle{thmstylethree}%

\begin{document}

\title[Machine Unlearning for Medical Imaging]{Machine Unlearning for Medical Imaging}





\author[1,3]{\fnm{Reza} \sur{Nasirigerdeh}}

\author[2]{\fnm{Nader} \sur{Razmi}}

\author[1,3,5]{\fnm{Julia A.} \sur{Schnabel}}

\author[3,4]{\fnm{Daniel} \sur{Rueckert}}

\author[1,3,4]{\fnm{Georgios} \sur{Kaissis}}




\affil[1]{\orgname{Helmholtz Munich}, \orgaddress{\city{Munich}, \country{Germany}}}

\affil[2]{\orgname{Independent Researcher}, \orgaddress{\city{Ardebil}, \country{Iran}}}

\affil[3]{\orgname{Technical University of Munich}, \orgaddress{\city{Munich}, \country{Germany}}}

\affil[4]{\orgname{Imperial College London}, \orgaddress{\city{London}, \country{United Kingdom}}}

\affil[5]{\orgname{King’s College London}, \orgaddress{\city{London}, \country{United Kingdom}}}


\abstract{\textit{Machine unlearning} is the process of removing the impact of a particular set of training samples from a pretrained model. It aims to fulfill the \enquote{right to be forgotten}, which grants the individuals such as patients the right to reconsider their contribution in models including medical imaging models. In this study, we evaluate the effectiveness (performance) and computational efficiency of different unlearning algorithms in \textit{medical imaging domain}. Our evaluations demonstrate that the considered unlearning algorithms perform well on the retain set (samples whose influence on the model is allowed to be retained) and forget set (samples whose contribution to the model should be eliminated), and show no bias against male or female samples. They, however, adversely impact the generalization of the model, especially for larger forget set sizes. Moreover, they might be biased against easy or hard samples, and need additional computational overhead for hyper-parameter tuning. In conclusion, machine unlearning seems promising for medical imaging, but the existing unlearning algorithms still needs further improvements to become more practical for medical applications.}




\maketitle

\section{Introduction}
\label{sec:intro}
\textit{Machine learning} (\textit{ML}) has considerably enhanced the capabilities of \textbf{\textit{medical image analysis}} \cite{dhawan2011-medical-image-analysis-1, duncan2000-medical-image-analysis-2} in various tasks including disease classification \cite{li2014-medical-classification-1, yadav2019-medical-classification-2} and semantic segmentation \cite{cai2020-medical-segmentation-1, wang2022-medical-segmentation-2}. ML models especially deep neural networks, however, rely on large-scale datasets \cite{lecun2015-deep-learning-1, goodfellow2016-deep-learning-2}, and they might leak privacy-sensitive information about specific samples in the dataset \cite{nasr2019-privacy-leakage1, shokri2017-privacy-leakage2, yeom2018-privacy-leakage3}. This can motivate the individuals such as patients to reconsider their contribution to such medical datasets and to the models trained on them. 

\textbf{\textit{Machine unlearning}} \cite{cao2015-machine-unlearning1, bourtoule2021-machine-unlearning2} is the process of eliminating the influence of given training samples from a trained machine learning model. Unlearning aims to fulfill the \enquote{right to be forgotten}, stipulated in Article 17 of \textit{General Data Protection Regulation} (\textit{GDPR}) \cite{voigt2017-gdpr}, which grants individuals the right to ask data holders such as hospitals and medical centers to have their data \enquote{forgotten}. \textit{Data forgetfulness} implies that data holders are obligated to \textit{unlearn} any model trained on the samples of given individuals to remove the contribution of the samples from the model. Moreover, they must delete those samples from their data records (known as \textit{data removal}). Machine unlearning might also have other applications in which data protection or privacy is not necessarily the major concern: eliminating the adverse impact of the noisy labels \cite{frenay2013-label-noise-survey, torkzadehmahani2023-confes} or harmful samples \cite{wehrli2022-bias-dl, fabbrizzi2022-bias-vis-data} on the model performance. Such applications (non privacy-focused) are interesting to companies to amend models by e.g. removing toxic content, but privacy and biases are more relevant to medicine, which have not yet been explored in prior works.

More formally, we assume there is a \textit{training set} on which the \textit{original model} has been already trained (\textit{pretrained model}), and a \textit{test set} on which the generalization of the model has been evaluated. We are also given (1) a \textit{forget set}, a subset of samples from the training set whose contribution to the pretrained model should be forgotten, and (2) a \textit{retain set}, another subset from the training set, whose contribution to  the pretrained model is still permitted to be retained. A given \textit{unlearning algorithm} takes the pretrained model, forget set, and retain set as arguments and returns the \textit{unlearned model} as output.  

Focusing on the \enquote{right to be forgotten} application of machine unlearning, the \textit{effectiveness} of different unlearning algorithms can be examined using the following criteria: On the retain set, the performance (e.g. accuracy) of the unlearned and pretrained models should be highly competitive; on the forget set, however, the unlearned model should underperform compared to the pretrained model. Optimally, the performance of the unlearned model on the forget set should not be much better than the performance of the pretrained model on a random sample from the population (i.e. train and test sets). Finally, the unlearned model should provide highly similar performance on the test set in comparison with the pretrained model, i.e. the unlearned model should still generalize well.

From the effectiveness perspective, the optimal algorithm for machine unlearning is \textit{exact unlearning}, where the pretrained model is first reinitialized with random weights, and then trained only on the retain set from scratch. Exact unlearning, however, incurs extremely high computational overhead, and consequently, it is impractical especially for very large models or frequent unlearning requests. \textit{Approximate unlearning} \cite{bourtoule2021-machine-unlearning2} algorithms address this challenge by conducting the unlearning process in a more computationally-efficient manner, but they might not be necessarily as performant as exact unlearning. Note that efficiency here refers to \textit{computational} efficiency, and by effectiveness, we mean the performance of the unlearned model on the retain, forget, and test sets. Given that, exact unlearning is effective but inefficient, and approximate unlearning algorithms, optimally, should be as effective as exact unlearning, but provide much more efficient unlearning process compared to exact unlearning.

Prior studies evaluated several approximate unlearning algorithms including \textit{random relabeling} of the forget set \cite{graves2021-relabling} or \textit{saliency unlearning} \cite{fan2023-salun}. These studies, however, are not representative for medical image applications due to the specifics of medical datasets such as the imbalanced nature of many medical datasets, the inclusion of multiple samples per patient in some datasets, etc. Consequently, a proof-of-concept for machine unlearning in the medical imaging domain is still missing in the literature.

In this paper, we investigate the effectiveness and efficiency of the existing state-of-the-art approximate unlearning algorithms for the \textit{medical imaging} on two large datasets, TissueMNIST (histopathology images) \cite{medmnistv1, medmnistv2}, and CheXpert (chest radiographs) \cite{irvin2019-chexpert}, using two convolutional network architectures (ResNet-18 and ResNet-50) \cite{he2016-deep-resnets}. Our evaluations demonstrate that 
\begin{itemize}
    \item The considered approximate unlearning algorithms achieve highly competitive performance on the retain and forget sets compared to exact unlearning.
    \item They underperform on the test set, implying that they negatively impact the generalization of the model, especially for bigger forget set sizes.
    \item They can be biased towards easy-to-classify or hard-to-classify samples, where they might provide significantly lower performance on either of the aforementioned group of samples.
    \item The algorithms show no bias towards male or females samples of the dataset.
    \item They need additional hyper-parameter tuning compared to exact unlearning, which adversely affect the efficiency of the algorithms.
\end{itemize}

Our results indicate a gap in current literature. Concretely, we argue that \textit{future approximate unlearning} algorithms for \textit{medical imaging} applications should (1) improve the generalization of the unlearned model compared to the state-of-the-art algorithms, and optimally, achieve highly comparable performance on the test set in comparison with exact unlearning, (2) provide unbiased performance on all group of samples including easy/hard samples, improving the fairness. In addition to that, future algorithms should minimize the additional computational overhead for hyper-parameter tuning, and ideally, obtain the desirable performance using the same hyper-parameter values as the original training.

\section{Results}
\label{sec:results}
To evaluate the performance of different approximate unlearning algorithms for medical image classification, we employ  TissueMNIST \cite{medmnistv1, medmnistv2} and CheXpert \cite{irvin2019-chexpert} as medical datasets, ResNet-18 and ResNet-50 \cite{he2016-deep-resnets} as models, and the \textit{area under ROC curve} (\textit{AUROC}) as our evaluation metric. Additionally, we leverage exact unlearning as our baseline, which indicates the optimal performance an approximate algorithm can achieve. Our evaluation is divided into three categories: (1) \textit{Forget set size based analysis}, which examines the effect of the forget set size on the performance of the unlearning algorithms, (2) \textit{per-class performance analysis}, which investigates how the unlearning algorithms impact the performance of the model on three representatives classes (easy-to-classify, intermediate-to-classify, and hard-to-classify), and (3) \textit{fairness analysis}, which separately examines the performance of the unlearning algorithms on the male and female samples to identify possible biases towards a specific group.

\vspace{1\baselineskip}
\noindent \textbf{Forget set size based analysis}. Tables \ref{tab:avg-tissue-mnist}-\ref{tab:avg-chexpert} list the \textit{average} AUROC values from different unlearning algorithms for three different sizes of the forget set ($5\%$, $15\%$, and $30\%$ of the training set size). As shown in the tables, on the \textit{forget set}, both \textit{random relabeling} and \textit{saliency unlearning} achieve highly competitive AUROC compared to exact unlearning for all three forget set sizes. They, however, underperform on the \textit{test set}, especially for the largest forget set size. On the \textit{retain set}, moreover, the algorithms provide very similar performance in comparison to exact unlearning for all considered forget set sizes, except forget set size of $30\%$ on \textit{TissueMNIST-ResNet-18}. 
\begin{table*}[!h]
    \centering
    \vskip -0.1in
    \caption{\textbf{Effect of forget set size} (\textit{TissueMNIST-ResNet-18}, \textit{Average} AUROC)}
    \vskip -0.1in
    \begin{minipage}{1.0\textwidth}
        \resizebox{\columnwidth}{!}{\begin{tabular}{l ccccccccccc}
            \toprule
              \multirow{2}{*}{} & \multicolumn{3}{c}{Forgetting=5\%} & &\multicolumn{3}{c}{Forgetting=15\%} & & \multicolumn{3}{c}{Forgetting=30\%} \\ \cline{2-4} \cline{6-8} \cline{10-12}
              Algorithm & Retain & Forget & Test & & Retain & Forget & Test & & Retain & Forget & Test \\
            \midrule
             Exact Unlearning & $97.88_{\pm0.10}$ & $91.80_{\pm0.03}$ & $92.00_{\pm0.16}$ & & $97.54_{\pm0.14}$ & $91.87_{\pm0.15}$ & $91.89_{\pm0.11}$ & & $97.51_{\pm0.28}$ & $91.52_{\pm0.20}$ & $91.50_{\pm0.17}$ \\
             Random Relabeling & $96.50_{\pm0.03}$ & $91.29_{\pm0.30}$ & $90.58_{\pm0.12}$ & & $96.20_{\pm0.21}$ & $91.67_{\pm0.43}$ & $90.55_{\pm0.38}$ & & $94.82_{\pm0.26}$ & $90.74_{\pm0.24}$ & $89.44_{\pm0.33}$ \\
             Saliency Unlearning & $96.68_{\pm0.18}$ & $91.50_{\pm0.24}$ & $90.58_{\pm0.10}$ & & $96.00_{\pm0.09}$ & $91.62_{\pm0.21}$ & $90.42_{\pm0.25}$ & & $95.04_{\pm0.32}$ & $91.12_{\pm0.29}$ & $89.78_{\pm0.22}$ \\
            \bottomrule
            \end{tabular}
        }
    \end{minipage}
\label{tab:avg-tissue-mnist}
\end{table*}
\begin{table*}[!h]
    \centering
    \vskip -0.1in
    \caption{\textbf{Effect of forget set size} (\textit{CheXpert-ResNet-50}, \textit{Average} AUROC)}
    \vskip -0.1in
    \begin{minipage}{1.0\textwidth}
        \resizebox{\columnwidth}{!}{\begin{tabular}{l ccccccccccc}
            \toprule
              \multirow{2}{*}{} & \multicolumn{3}{c}{Forgetting=5\%} & &\multicolumn{3}{c}{Forgetting=15\%} & & \multicolumn{3}{c}{Forgetting=30\%} \\ \cline{2-4} \cline{6-8} \cline{10-12}
              Algorithm & Retain & Forget & Test & & Retain & Forget & Test & & Retain & Forget & Test \\
            \midrule
             Exact Unlearning & $80.19_{\pm0.13}$ & $78.36_{\pm0.07}$ & $77.51_{\pm0.05}$ & & $80.10_{\pm0.07}$ & $77.66_{\pm0.02}$ & $77.36_{\pm0.09}$ & & $79.71_{\pm0.21}$ & $77.56_{\pm0.12}$ & $77.10_{\pm0.07}$ \\
             Random Relabeling & $80.85_{\pm0.19}$ & $78.13_{\pm0.16}$ & $77.15_{\pm0.07}$ & & $80.02_{\pm0.29}$ & $77.01_{\pm0.32}$ & $76.63_{\pm0.44}$ & & $79.16_{\pm0.14}$ & $76.81_{\pm0.07}$ & $76.19_{\pm0.13}$ \\
             Saliency Unlearning & $81.10_{\pm0.31}$ & $78.17_{\pm0.24}$ & $77.33_{\pm0.11}$ & & $80.55_{\pm0.10}$ & $77.40_{\pm0.14}$ & $76.90_{\pm0.18}$ & & $79.32_{\pm0.05}$ & $77.02_{\pm0.07}$ & $76.12_{\pm0.30}$ \\
            \bottomrule
            \end{tabular}
        }
    \end{minipage}
\label{tab:avg-chexpert}
\end{table*}

\vspace{1\baselineskip}
\noindent \textbf{Per-class performance analysis}. Tabels \ref{tab:per-class-tissue-mnist}-\ref{tab:per-class-chexpert} show the AUROC values for three chosen classes, i.e. easy-to-classify, intermediate-to-classify, and hard-to-classify. The difficulty of a given class is determined based on the performance of the pretrained model on the samples of that class from the test set, where lower model performance on a class implies higher difficulty of the class. According to the tables, on the \textit{retain set}, the \textit{random relabeling} and \textit{saliency unlearning} algorithms provide highly similar performance compared to exact unlearning for \textit{easy/intermediate} classes on both TissueMNIST-ResNet-18 and CheXpert-ResNet-50 case studies. This is also the case for the \textit{hard} class on CheXpert-ResNet-50. On TissueMNIST-ResNet-18, however, they significantly impact the performance on the \textit{retain set} for the \textit{hard} class.

On the \textit{forget set} and \textit{test set} of TissueMNIST, the \textit{random relabeling} and \textit{saliency unlearning} algorithms significantly underperform on the \textit{easy} and \textit{hard} classes, respectively. For the other cases, however, they provide highly comparable or slightly lower AUROC values on the \textit{forget} and \textit{test} sets compared to exact unlearning. 
\begin{table*}[!h]
    \centering
    \vskip -0.1in
    \caption{\textbf{Per-class performance} (\textit{TissueMNIST-ResNet-18}, AUROC, Forgetting=15\%)}
    \vskip -0.1in
    \begin{minipage}{1.0\textwidth}
        \resizebox{\columnwidth}{!}{\begin{tabular}{l ccccccccccc}
            \toprule
              \multirow{2}{*}{} & \multicolumn{3}{c}{Class=Easy} & &\multicolumn{3}{c}{Class=Intermediate} & & \multicolumn{3}{c}{Class=Hard} \\ \cline{2-4} \cline{6-8} \cline{10-12}
              Algorithm & Retain & Forget & Test & & Retain & Forget & Test & & Retain & Forget & Test \\
            \midrule
             Exact Unlearning & $98.64_{\pm0.15}$ & $95.82_{\pm0.11}$ & $96.23_{\pm0.09}$ & & $97.43_{\pm0.16}$ & $92.15_{\pm0.17}$ & $92.13_{\pm0.15}$ & & $95.18_{\pm0.41}$ & $84.19_{\pm0.58}$ & $83.90_{\pm0.28}$ \\
             Random Relabeling & $97.33_{\pm0.20}$ & $92.19_{\pm1.17}$ & $92.50_{\pm1.27}$ & & $96.11_{\pm0.18}$ & $91.79_{\pm0.63}$ & $90.81_{\pm0.55}$ & & $92.68_{\pm0.55}$ & $84.81_{\pm0.46}$ & $81.64_{\pm0.21}$ \\
             Saliency Unlearning & $97.95_{\pm0.19}$ & $95.3_{\pm0.33}$ & $94.88_{\pm0.24}$ & & $95.81_{\pm0.20}$ & $91.94_{\pm0.35}$ & $90.72_{\pm0.24}$ & & $92.49_{\pm0.75}$ & $82.79_{\pm1.88}$ & $79.44_{\pm2.87}$ \\
            \bottomrule
            \end{tabular}
        }
    \end{minipage}
\label{tab:per-class-tissue-mnist}
\end{table*}

\begin{table*}[!h]
    \centering
    \vskip -0.1in
    \caption{\textbf{Per-class performance}  (\textit{CheXpert-ResNet-50}, AUROC, Forgetting=15\%)}
    \vskip -0.1in
    \begin{minipage}{1.0\textwidth}
        \resizebox{\columnwidth}{!}{\begin{tabular}{l ccccccccccc}
            \toprule
              \multirow{2}{*}{} & \multicolumn{3}{c}{Class=Easy} & &\multicolumn{3}{c}{Class=Intermediate} & & \multicolumn{3}{c}{Class=Hard} \\ \cline{2-4} \cline{6-8} \cline{10-12}
              Algorithm & Retain & Forget & Test & & Retain & Forget & Test & & Retain & Forget & Test \\
            \midrule
             Exact Unlearning & $87.95_{\pm0.05}$ & $86.39_{\pm0.09}$ & $86.56_{\pm0.08}$ & & $86.22_{\pm0.15}$ & $83.11_{\pm0.09}$ & $82.45_{\pm0.12}$ & & $69.54_{\pm0.14}$ & $66.99_{\pm0.09}$ & $67.10_{\pm0.21}$ \\
             Random Relabeling & $88.01_{\pm0.26}$ & $86.22_{\pm0.17}$ & $86.20_{\pm0.55}$ & & $85.44_{\pm0.49}$ & $82.35_{\pm0.50}$ & $80.67_{\pm0.76}$ & & $69.77_{\pm0.32}$ & $65.95_{\pm0.45}$ & $66.88_{\pm0.44}$ \\
             Saliency Unlearning & $88.36_{\pm0.19}$ & $86.40_{\pm0.08}$ & $86.31_{\pm0.03}$ & & $86.08_{\pm0.00}$ & $82.74_{\pm0.31}$ & $80.95_{\pm0.51}$ & & $70.25_{\pm0.23}$ & $66.59_{\pm0.18}$ & $67.47_{\pm0.14}$ \\
            \bottomrule
            \end{tabular}
        }
    \end{minipage}
\label{tab:per-class-chexpert}
\end{table*}
\begin{table*}[!h]
    \centering
    \vskip -0.1in
    \caption{\textbf{Fairness analysis} (\textit{CheXpert-ResNet-50}, AUROC, Forgetting=15\%)}
    \vskip -0.1in
    \begin{minipage}{1.0\textwidth}
        \resizebox{\columnwidth}{!}{\begin{tabular}{l ccccccccccc}
            \toprule
              \multirow{2}{*}{} & \multicolumn{2}{c}{Retain} & &\multicolumn{2}{c}{Forget} & & \multicolumn{2}{c}{Test} \\ \cline{2-3} \cline{5-6} \cline{8-9}
             Algorithm & Male & Female & & Male & Female & & Male & Female \\
            \midrule
                Exact Unlearning & $80.14_{\pm0.07}$ & $80.03_{\pm0.06}$ & & $77.80_{\pm0.06}$ & $77.47_{\pm0.05}$ & &  $77.24_{\pm0.09}$ & $77.43_{\pm0.11}$ \\
                Random Relabeling & $80.11_{\pm0.21}$ & $79.90_{\pm0.40}$ & & $77.16_{\pm0.21}$ & $76.79_{\pm0.46}$ & & $76.57_{\pm0.23}$ & $76.61_{\pm0.76}$ \\
                Saliency Unlearning & $80.59_{\pm0.10}$ & $80.51_{\pm0.08}$ & & $77.51_{\pm0.11}$ & $77.26_{\pm0.19}$ & & $76.85_{\pm0.09}$ & $76.90_{\pm0.35}$ \\
            \bottomrule
            \end{tabular}
        }
    \end{minipage}
\label{tab:fairness-chexpert}
\end{table*}

\vspace{1\baselineskip}
\noindent \textbf{Fairness analysis}. Table \ref{tab:fairness-chexpert} lists the average AUROC values from different learning algorithms on the male and female samples of the CheXpert dataset. As indicated in the table, the performance of each considered unlearning algorithm on male samples is highly competitive to that on the female samples. This implies that none of the unlearning algorithms has bias towards the male or female group in the dataset.

\section{Methods}
\label{sec:methods}
We provide a brief overview on the unlearning algorithms, datasets, training procedure, and unlearning process employed in the experiments.

\subsection{Unlearning algorithms}
As mentioned before, unlearning algorithms can be divided into \textit{exact unlearning} and \textit{approximate unlearning}. Although exact unlearning is optimal from the performance perspective, it is highly inefficient from the computational-efficiency aspect. Approximate unlearning, on the other hand, aims to incur very low computational overhead, but to achieve highly competitive performance compared to exact unlearning. In the following, we briefly describe exact unlearning, and the \textit{random relabeling} and \textit{saliency unlearning} approximate algorithms used in our evaluations.

\vspace{1\baselineskip}
\noindent \textbf{Exact unlearning.} This algorithm randomly initializes the weights of the pretrained model, and trains the model from scratch on the retain set. The algorithm follows the same training procedure (e.g. learning rate, number of epochs, etc) employed to train the original model. 

\vspace{1\baselineskip}
\noindent \textbf{Random relabeling.} The algorithm assigns a random label to the samples of the forget set to obtain the noisy forget set, creates a new training set by combining the retain set and the noisy forget set, and fine-tunes the pretrained model on the new training set. The hyper-parameters for the fine-tuning procedure might be very different from those used in the training process of the original model.  

\vspace{1\baselineskip}
\noindent \textbf{Saliency unlearning}. This algorithm first computes the gradient value corresponding to each parameter of the pretrained model on the \textit{forget set}. If the absolute value of the gradient is greater than a pre-specified threshold, then mask of 1 is allocated to the corresponding model parameter; otherwise, its mask is set to 0. Next, it relabels the samples of the forget set, and creates a new training set with the retain set and noisy forget set combined, exactly the same as the random relabeling algorithm. During unlearning, however, it only updates the model parameters with mask of 1, and freezes the other model parameters.

\subsection{Datasets}
\vspace{1\baselineskip}
\noindent \textbf{TissueMNIST}. This dataset comprises 165,466 train images, 23,640 validation images, and 47,280 test images from 8 classes. In our evaluations, we employ only 50,000 of the train images. All images are of shape 128$\times$128 pixels. The input pixel values of the images are divided by 255 to lie in range [0, 1].

\vspace{1\baselineskip}
    \noindent \textbf{CheXpert}. The dataset consists of 223,414 train images, out of which we use 5,285 and 5,025 images for validation and test, respectively. Splitting is performed based on patient IDs, where samples of a given patient can belong only to one of the train/validation/test sets. The images are resized from their original size of 320$\times$390 pixels to 224$\times$224 pixels. We use the \enquote{U-one policy} \cite{irvin2019-chexpert}, where the unknown labels are assumed to be positive labels, and consider \enquote{Atelectasis}, \enquote{Cardiomegaly}, \enquote{Consolidation}, \enquote{Edema}, \enquote{Pleural Effusion} as labels \cite{irvin2019-chexpert}. Note that each label can have a binary value (1 for positive class and 0 for negative class). Similar to TissueMNIST, the input pixel values are divided by 255.

\subsection{Training}
We train ResNet-18 and ResNet-50 models on the TissueMNIST and CheXpert datasets, respectively. We adopt the the original implementations of the models from the PyTorch library \cite{Paszke2019-pytorch}, which employ BatchNorm \cite{ioffe2015-batch-norm} as the normalization layer.

\vspace{1\baselineskip}
\noindent \textbf{Original training}. We train the randomly initialized model on the train set for 6 epochs with batch size of 32, the Adam optimizer \cite{kingma2014-adam-optimizer}, and initial learning rate of 0.001, which is gradually decayed by factor of 0.1 using Cosine Annealing scheduler \cite{loshchilov2016-cosine-annealing}. The loss functions are cross-entropy and binary cross-entropy with logits for the TissueMNIST-ResNet-18 and CheXpert-ResNet-50 case studies, respectively. We use the weights of the model in the last epoch as the pretrained model.

\vspace{1\baselineskip}
\noindent \textbf{Exact unlearning}. We employ the same training procedure as the original training, except that the underlying dataset is the retain set, consisting of around 95\%, 85\%, and 70\% of the samples from the train set depending on the corresponding forget set size, which can be about 5\%, 15\%, and 30\% of the train set size, respectively.  

\vspace{1\baselineskip}
\noindent \textbf{Approximate unlearning}. For both \textit{random relabeling} and \textit{saliency unlearning} algorithms, we unlearn the pretrained model for 2 epochs. We perform hyper-parameter tuning over different learning rate and gradient threshold (only for saliency unlearning) values to obtain the most desirable result, which is determined based on the average AUROC on the forget set compared to exact unlearning.



\section{Discussion}
\label{sec:discussion}
We conduct the proof-of-concept of machine unlearning in medical imaging by evaluating the performance of several approximate unlearning algorithms including \textit{random relabeling} and \textit{saliency unlearning} on medical image classification using the TissueMNIST and CheXpert medical datasets. Our results indicate that the considered algorithms work well on the retain and forget sets in comparison to exact unlearning. Moreover, they show no bias against male or female samples in the dataset. The algorithms, however, adversely impact the generalization of the model (i.e. performance on the test set), particularly for larger forget set sizes. They, additionally, can be biased against \textit{easy} or \textit{hard} classes. For instance, the \textit{random relabeling} algorithm significantly underperform on the \textit{easy} class of the TissueMNIST dataset. The \textit{saliency unlearning} algorithm, on the other hand, provides considerably lower performance on the \textit{hard} class.

Additionally, the difficulty of the overall task seems to effect the unlearning performance. For instance, the unlearning algorithms achieve lower performance on TissueMNIST in comparison with CheXpert. This can be due to fact that we employed the binary-class, multi-label (5 labels) version of CheXpert, whereas TissueMNIST is a multi-class dataset (8 classes). Given that both \textit{random relabeling} and \textit{saliency unlearning} algorithms assign a random class to the samples of the forget set, the samples of TissueMNIST can be allocated a more diverse range of random classes compared to CheXpert.

Hyper-parameter tuning is an essential part of any machine learning or unlearning process. One of the positive characteristics of \textit{exact unlearning} is that the algorithm incurs no additional overhead for hyper-parameter tuning. That is, we can employ the same hyper-parameter values (e.g. learning rate) and training procedure as the original training used to obtain the pretrained model. The approximate unlearning algorithms, on the other hand, requires additional computation to perform hyper-parameter tuning. For instance, the learning rate should be tuned for the \textit{random relabeling} algorithm, or the \textit{saliency unlearning} algorithm introduces a new hyper-parameter (i.e. gradient threshold), which needs to be tuned to achieve the desirable performance.

In our evaluations, we focus on ResNets, one of the most popular architectures in medical image classification. Moreover, we employ a multi-class, single-label dataset (TissueMNIST) and a binary-class, multi-label dataset (CheXpert) in our experiments. Additionally, we use AUROC as our performance metric. Investigating the performance of approximate unlearning algorithms using other architectures such as vision transformers \cite{dosovitskiy2020-vit}, conducting the evaluations on a multi-class, multi-label medical image dataset, and capitalizing on other metrics such as membership inference attack (MIA) success rate to investigate the effectiveness of the approximate unlearning algorithms \cite{hayes2024-mia-unlearning} are all interesting directions for future studies.

In conclusion, we observe that machine unlearning is promising for medical image classification. However, the performance of the existing unlearning algorithms needs further improvements in terms of the generalization of the unlearned model, bias towards/against a specific group of samples, or additional computational overhead for hyper-parameter tuning to become more practical in the medical imaging domain.

\backmatter

\bibliography{sn-article.bib}

\end{document}